%% file: main.tex
\documentclass{pinepaper}

\usepackage{amssymb,amsfonts}
\usepackage{algorithmic}
\usepackage{algorithm}
\usepackage{textcomp}
\usepackage{etoolbox,array}
\usepackage{siunitx}
\usepackage{multirow}
\usepackage{needspace}
\usepackage{adjustbox}

\newsavebox{\contactResultantPanel}
\newsavebox{\contactForcesPanel}
\newsavebox{\contactResidualsPanel}

\title{DexTeleop-0: Force-Aware Bimanual Dexterous Teleoperation with Ego-Centric Perception towards Shared Autonomy}

\author[1]{Haichao Liu}
\author[1]{Yuyao Jiang}
\author[2]{Hyunsun Park}
\author[1]{Yuanjiang Xue}
\author[1]{\authorcr Mingze Sun}
\author[1]{Ziwei Liu}
\author[1]{Ziwei Wang\textsuperscript{\textdagger}}
\affil[1]{Nanyang Technological University, Singapore}
\affil[2]{OOJU, USA}

\date{\textsuperscript{\textdagger}Corresponding author.}

\papervenue{Preprint}
\correspondence{\href{mailto:ziwei.wang@ntu.edu.sg}{ziwei.wang@ntu.edu.sg}}

\begin{document}
\makepinetitle

\begin{pineabstract}
Fine-grained, bimanual dexterous manipulation remains a foundational challenge in robotics. Embodiment gaps hinder accurate kinematic mapping, \textcolor{black}{and visual tracking alone does not regulate contact forces.} To address these limitations, we propose a tactile-driven adaptation strategy designed to enable fine-grained manipulation on top of teleoperation pipelines. Instantiated within our bimanual dexterous framework, DexTeleop-0, this strategy introduces a real-time optimization loop that \textcolor{black}{adds robot-side contact corrections to human tracking commands using fingertip feedback.} By estimating contact positions and forces, the system dynamically computes localized corrections using \textcolor{black}{a contact-stiffness approximation and the kinematic Jacobian, with per-contact vector force tracking, soft predicted-contact-moment regularization, and bounded hand-joint residuals.} We evaluate this tactile-driven adaptation strategy across both simulated environments and real-world hardware. \textcolor{black}{Controlled simulation replay yields final-stage success rates of 91\% for Ball Assembly and 62\% for Stir in Cup, versus 88\% and 49\% for force tracking alone. Fixed-input replay on two hand types, seven-operator teleoperation, and external-force traces further assess contact assistance. The open-sourced project page can be found at \url{https://henryhcliu.github.io/dexteleop-0}.
}
\end{pineabstract}


\section{Introduction}
\label{sec:introduction}

Multi-fingered robotic hands offer the structural versatility required to execute contact-rich, fine-grained manipulation tasks beyond the capabilities of conventional parallel grippers. Tactile sensing provides critical physical insights into localized pressure profiles, transient slip conditions, and multi-point contact dynamics~\cite{yuan2017gelsight,suresh2024neuralfeels}. Bimanual configurations allow robots to stabilize non-prehensile objects and orchestrate highly synchronized dual-arm assemblies. Human behavioral data include imitation-learning demonstrations, passive human videos, and egocentric tracking datasets~\cite{qin2022dexmv,chi2024umi,wang2024dexcap}. \textcolor{black}{Teleoperation provides one route to collecting demonstrations, whose usefulness depends on the robot executing the intended contacts despite differences in embodiment.}

Autonomous dexterous manipulation has advanced through learned grasping~\cite{xu2023unidexgrasp,fang2025anydexgrasp}, task planning and motion control~\cite{liu2025robodexvlm}, and generalizable controllers and manipulation priors~\cite{yin2025dexteritygen,caggiano2023myodex}. Demonstration augmentation~\cite{wang2024cyberdemo}, vision-language-action models~\cite{zhong2026dexgraspvla}, and visuomotor imitation policies~\cite{chi2023diffusionpolicy,ze2024dp3} offer complementary routes to broader task coverage. For demonstration-based approaches, the quality and diversity of recorded interactions remain important. \textcolor{black}{This motivates teleoperation interfaces that preserve operator intent while assisting with contact errors, without requiring a learned policy for each manipulation task.}

Teleoperation interfaces include leader--follower systems such as GELLO~\cite{wu2023gello}, ALOHA~\cite{zhao2023aloha}, and ALOHA~2~\cite{aloha2team2024aloha2}. To capture highly expressive hand motions, researchers explored mechanical and visual-exoskeleton interfaces~\cite{yang2025ace} or traditional input options like multi-axis joystick controllers~\cite{dhat2024using3dmice}. Their suitability depends on the required coordination and operator workload. Vision-based retargeting systems observe the human hand using standalone cameras to map coordinates onto robotic alternatives~\cite{sivakumar2022telekinesis,qin2022one2many,qin2023anyteleop}. This avoids a mechanically matched leader but remains sensitive to visibility and tracking accuracy. Headset-based systems extend this paradigm with egocentric tracking and visual feedback~\cite{ding2024bunnyvisionpro,cheng2024opentelevision}. DexTeleop-0 builds on this established tracking and retargeting front end.

\begin{figure}[!htb]
  \centering
  \includegraphics[width=.68\columnwidth]{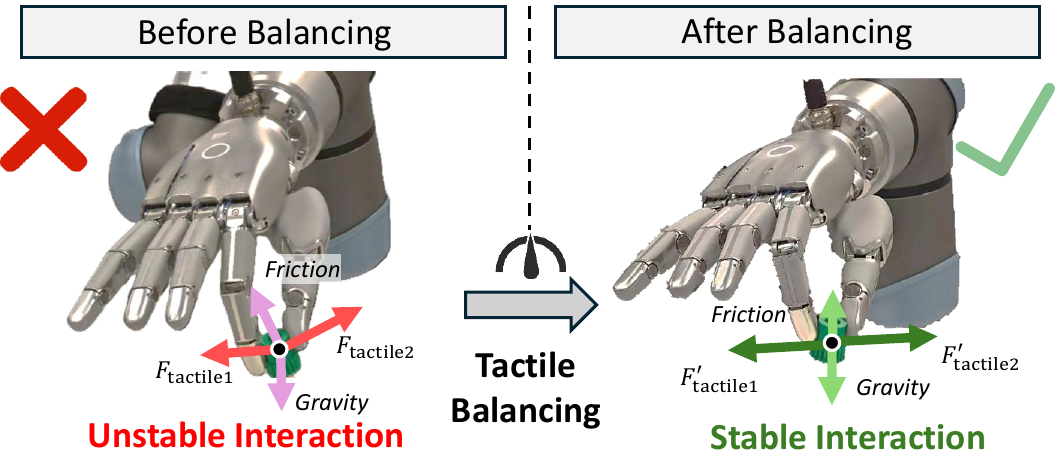}
  \caption{Contact correction through posture adjustment. \textcolor{black}{Changes in finger posture modify individual contact forces and their combined moment.}}
  \label{fig:tactile_balancing_principle}
\end{figure}

Despite this potential, acquiring high-quality bimanual dexterous data presents a significant bottleneck. A major underlying issue is the inherent \textit{embodiment gap} between human hands and multi-fingered robotic topologies, which \textcolor{black}{can leave errors in retargeted contact positions and forces. In an interface without force reflection, the operator must infer these errors from visual observation. Robot-side contact correction provides an additional feedback path: the human supplies nominal motion, while local sensing adjusts the robot command during contact. The aim is to assist execution while retaining the operator's overall manipulation strategy.}

Tactile information has supported contact-rich policy learning~\cite{xue2025reactivediffusion,wu2024tacdiffusion}, in-hand visual-tactile perception and manipulation~\cite{yuan2024robotsynesthesia,suresh2024neuralfeels}, and visual-tactile pretraining across tasks~\cite{qivlsual2026}. \textcolor{black}{Force-aware systems also use different feedback paths. GELLO + Force~\cite{sujit2025gello_force} reflects follower interaction forces through an actuated physical leader. DexForce~\cite{dexforce2025} extracts force-informed fingertip position targets from kinesthetic demonstrations and tracks them with Cartesian impedance control. It also evaluates policy learning. These approaches motivate the use of contact information but differ from online robot-hand correction around visually tracked targets.}

To bridge this gap, we introduce a \textbf{tactile-driven adaptation strategy} for bimanual dexterous teleoperation, implemented within our framework named \textbf{DexTeleop-0}. \textcolor{black}{Fingertip forces and contact positions enter an explicit local stiffness approximation coupled to the kinematic Jacobian. The objective retains separate vector force errors at each contact and adds soft predicted-contact-moment regularization, subject to bounded hand-joint residuals. Arm commands remain nominal, and each hand is optimized independently. As illustrated in Fig.~\ref{fig:tactile_balancing_principle}, posture corrections modify individual contact forces and their combined moment. The formulation couples fingertip regulation and multi-contact coordination around visually tracked targets, providing robot-side contact assistance while preserving the operator's nominal motion.}

Our primary contributions are as follows:
\begin{itemize}
\item \textcolor{black}{A fingertip-force-informed optimization formulation combining per-contact vector tracking, local stiffness, predicted-moment regularization, and bounded joint residuals around teleoperation targets.}
\item \textcolor{black}{A first-order contact interaction model using contact positions, stiffness approximations, and kinematic Jacobians to predict fingertip-force and multi-contact moment changes induced by hand-joint corrections.}
\item \textcolor{black}{A robot-side shared-autonomy framework integrating egocentric tracking and retargeting with fingertip-feedback control on a high-DoF bimanual platform, refining operator-commanded hand motions during contact.}
\item \textcolor{black}{Experimental evidence from simulation and real-world manipulation demonstrating improved task completion over nominal retargeting, with task-dependent force and timing tradeoffs and benefits over force tracking alone.}
\end{itemize}

\section{Methodology}\label{sec:method}
\subsection{Problem Formulation}
\label{subsec:problem_formulation}\label{subsec:interface}
\begin{figure}[!htb]
  \centering
  \includegraphics[width=.74\columnwidth]{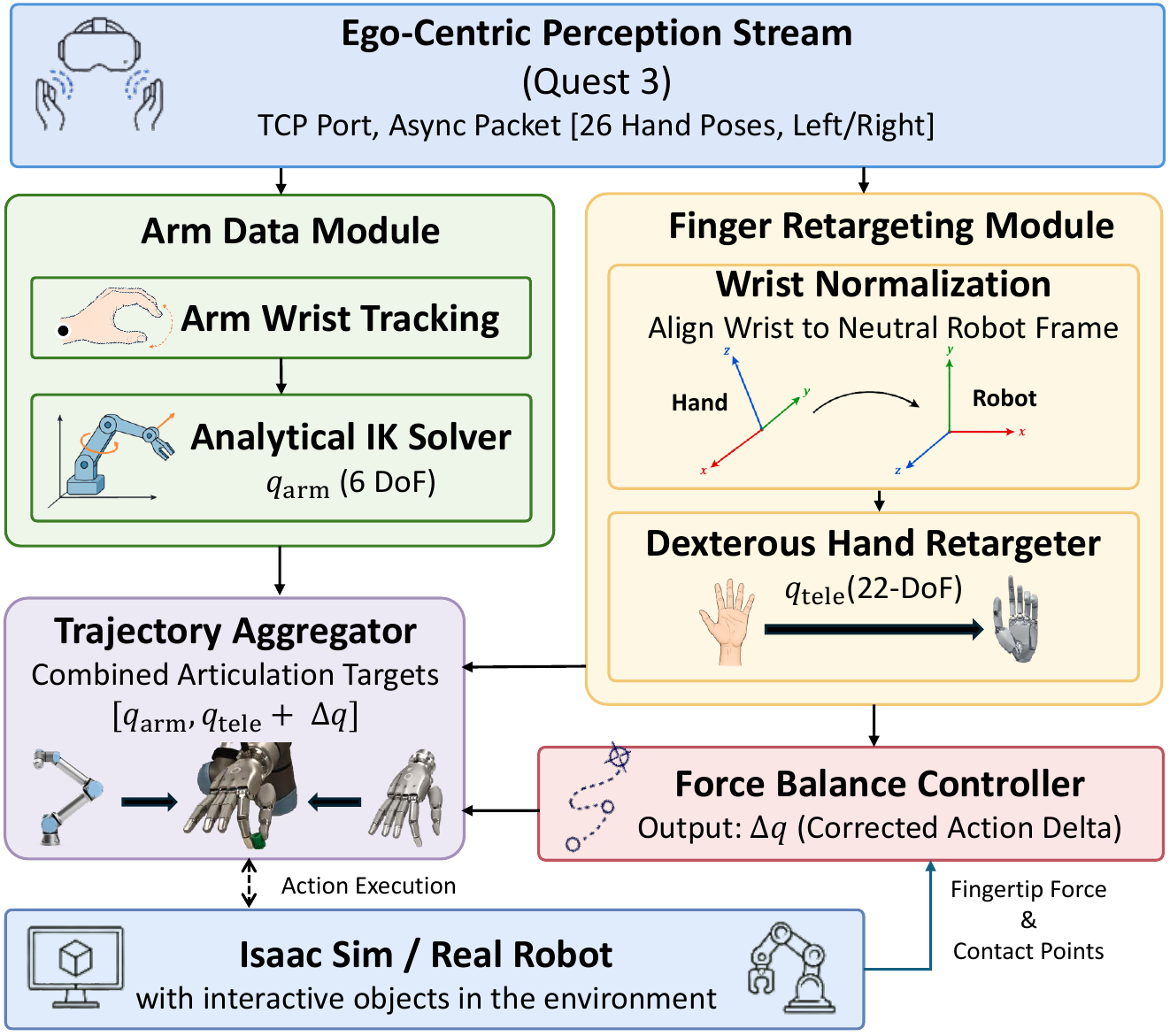}
  \caption{DexTeleop-0 architecture: egocentric tracking, arm IK, hand retargeting, and robot-side contact correction. \textcolor{black}{Residuals modify hand targets, with each hand solved independently.}}
  \label{fig:system_architecture}
\end{figure}

The system maps VR wrist poses to nominal arm commands $q_a$ through inverse kinematics and tracked hand keypoints to nominal hand commands $q_{\rm nom}\in\mathbb R^n$ through vector retargeting. \textcolor{black}{As shown in Fig.~\ref{fig:system_architecture}, the residual controller adds $\Delta q\in\mathbb R^n$ to each hand target, while retaining the arm command: $q^*=[q_a^T,(q_{\rm nom}+\Delta q)^T]^T$. Each hand is optimized separately. The operator receives visual feedback, while fingertip feedback closes an autonomous robot-side loop without haptic force reflection. Robot-specific kinematics, limits, contact geometry, and sensing adapters define the controller inputs.}

\subsection{Dexterous Hand Retargeting}
\label{subsec:dexterous_hand_retargeting}
The nominal hand command matches tracked human finger-to-finger and finger-to-palm vectors to robot vectors $v_k(q)=p_{{\rm task},k}(q)-p_{{\rm origin},k}(q)$. The vector matching and project-and-escape mechanism follow DexPilot~\cite{Handa2020DexPilot}. For selected tip pairs, this hysteresis replaces the reference length by a configured contact distance below a projection threshold, restores the tracked vector above an escape threshold, and retains the previous mode between thresholds. Projected pairs receive higher weights $w_k$. The nominal hand target solves
\begingroup
\begin{equation}
\begin{aligned}
q_{\rm nom}=\arg\min_q\quad &\frac1K\sum_{k=1}^K w_k\mathrm H_\eta
 \bigl(\|v_k(q)-v_k^{\rm ref}\|_2\bigr)\\
 &+\lambda_{\rm ret}\|q-q^{\rm last}\|_2^2,\\
\text{s.t.}\quad &q_{\min}\leq q\leq q_{\max}.
\end{aligned}
\label{eq:retarget_objective}
\end{equation}
\endgroup
Here $K$ is the number of vector pairs, $\mathrm H_\eta$ is the Huber loss, $\lambda_{\rm ret}$ weights temporal regularization, and $q^{\rm last}$ is the preceding target. SLSQP solves this constrained nonlinear problem using kinematic-Jacobian gradients. \textcolor{black}{Tracking, IK, and retargeting provide the nominal motion targets for the contact-residual controller below.}

\subsection{Contact-Rich Force-Aware Shared Autonomy}
\label{subsec:contact_rich_force_aware_shared_autonomy}
\textcolor{black}{The residual problem coordinates joint corrections using per-contact kinematics, vector force errors, and a soft predicted-moment cost. It operates online around nominal teleoperation targets, differing from independent finger-level Linear adjustment and DexForce's Cartesian target replay. Algorithm~\ref{alg:residual} summarizes the procedure.}

\subsubsection{Tactile Sensing and Contact State Estimation}\label{subsec:sensing}
\textcolor{black}{Contact feedback comprises a contact-position estimate $p_i$, a measured or estimated force vector $f_i$, and a confidence $c_i\in[0,1]$ for each valid contact $i\in\mathcal C$. These quantities are expressed in the hand-base frame with a hand-on-object force convention. The adapter determines contact validity before the force and moment objectives are assembled. Contact moments are computed from the position and force estimates.}

\textcolor{black}{For \textit{Sharpa Wave}, tactile-image contact locations are mapped to finger geometry, with a geometric position fallback, and associated with the available force-vector channels. Contact locations and finger kinematics supply $p_i$, while the force vector supplies $f_i$ at that position.
}
\textcolor{black}{For \textit{Wuji Hand 2}, which lacks direct fingertip force sensing, contact estimation uses joint position, velocity, and current feedback. For each four-joint finger, empirical joint torques are $\hat\tau_q=k_I\odot(I-I_b)$. URDF kinematics transform a configured pad point and normal $n_i$ to the hand-base frame. Under a quasi-static, single-contact, normal-force approximation, $\hat\tau_q\approx a_iF_{n,i}$, where $a_i=J_{i,f}^Tn_i$ and $J_{i,f}\in\mathbb R^{3\times4}$ is the finger-restricted contact Jacobian. Damped nonnegative least squares gives}
\begingroup\color{black}
\begin{equation}
\hat F_{n,i}=\max\left(0,\frac{a_i^T\hat\tau_q}{a_i^Ta_i+\lambda_n^2}\right),
\qquad \hat f_i=\hat F_{n,i}n_i.
\label{eq:wuji_force}
\end{equation}
\endgroup
\textcolor{black}{Defaults are $k_I=0.1$ N\,m/A, $I_b=0$ A, and $\lambda_n=0.001$ m.
Its contact model uses $K_i=k_i n_in_i^T$.
PhysX supplies contact positions and forces in simulation. These sensing adapters map the two hand configurations and simulation feedback to the common variables $p_i$, $f_i$, and $c_i$ used by the controller.}

\begin{algorithm}[!t]
\caption{\textcolor{black}{Robot-side residual contact control (per hand)}}
\label{alg:residual}
\begingroup\color{black}\small
\begin{algorithmic}[1]
\REQUIRE Nominal target $q_{\rm nom}$, previous residual $\Delta q_{\rm prev}$, joint/contact feedback, kinematics, stiffness, targets, weights, and bounds
\STATE Estimate contacts and forces, then transform to the hand-base frame with hand-on-object force convention
\STATE Reject invalid or non-contact feedback and form $\mathcal C$ and confidences $c_i$
\STATE Evaluate $p_i,J_i$ at $q_{\rm nom}$ and set $A_i=K_iJ_i$ and $f_i^{\rm tar}=\operatorname{clip}(f_i,l_i,u_i)$
\STATE Construct $L_F$ from stacked per-contact errors (\ref{eq:local_cost})
\IF{sufficient valid contacts for moment regularization}
\STATE Compute contact centroid, moment error, and $B_\tau$ (\ref{eq:resultants})--(\ref{eq:torque_jacobian})
\ELSE
\STATE Disable moment cost ($w_\tau=0$)
\ENDIF
\STATE Solve residual QP (\ref{eq:residual_qp}) with bounds (\ref{eq:bounds})
\STATE Validate $\Delta q$ and select a configured valid fallback residual on failure
\ENSURE Corrected hand command $q_{\rm cmd}=q_{\rm nom}+\Delta q$ and nominal arm command
\end{algorithmic}
\endgroup
\end{algorithm}

\input{revision_shared/controller_r1.tex}

\section{Experimental Results and Analysis}\label{sec:evaluation}
\subsection{Experimental Setup and Metrics}
\label{subsec:experimental_setup_and_metrics}\label{subsec:setup}
\subsubsection{System Architecture and Hardware Configuration}
\begin{figure}[!htb]
    \centering
    \includegraphics[width=.8\linewidth]{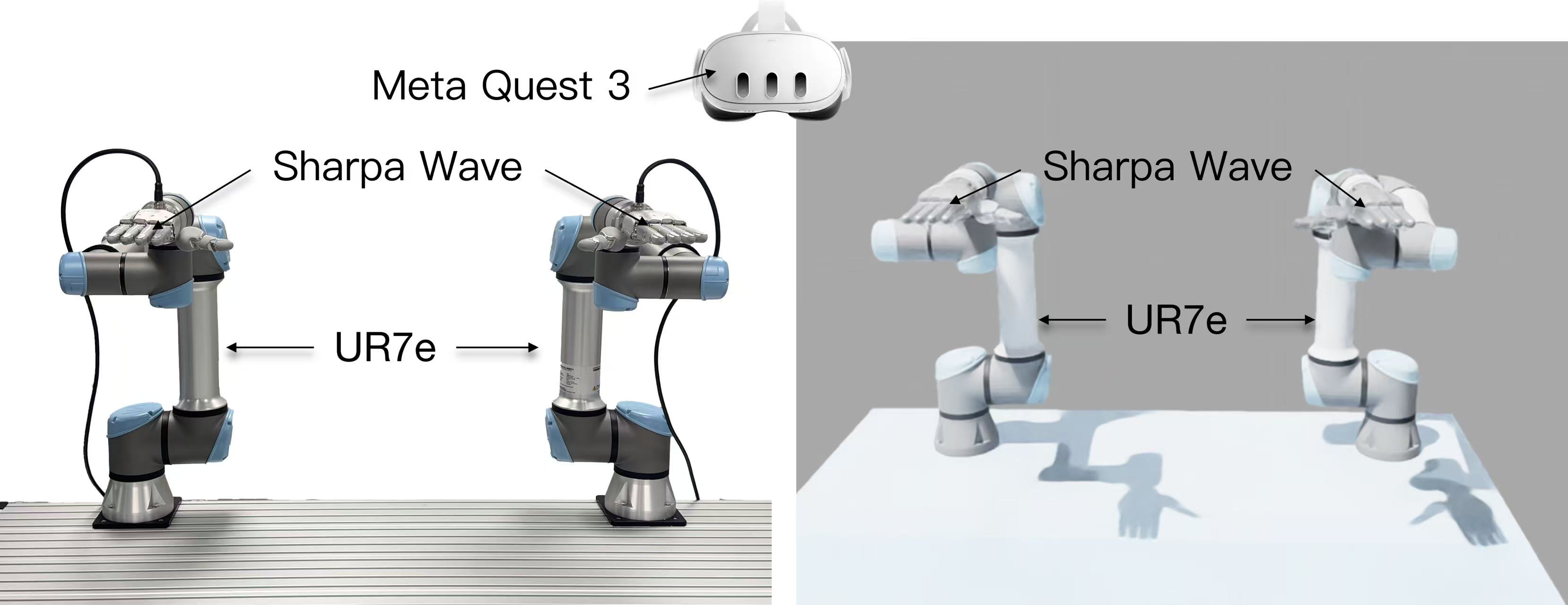}
    \caption{Physical bimanual platform (left) and IsaacSim counterpart (right), both using Meta Quest 3 input.}
    \label{fig:sim_real_align}
\end{figure}

The platform in Fig.~\ref{fig:sim_real_align} combines Meta Quest 3 tracking, two 6-DoF UR7e arms, two 22-DoF Sharpa Wave hands, and \textcolor{black}{IsaacSim 4.5. Additional experiment uses Wuji Hand 2 with 20 joints per hand. Simulation and hardware provide complementary evaluations of the residual controller. The controller updates at 30 Hz and uses a moment tolerance of 10 N\,m (10 $\mathrm N_{\rm est}$\,m for Wuji) with at least two valid contacts. The code and detailed parameter settings will be made publicly available upon acceptance.}

\subsubsection{Baselines and Ablation}
\textcolor{black}{The baselines are \textbf{No Residual}, which tracks retargeted positions, \textbf{Linear}, which independently adjusts each finger using a linear force-feedback rule, and \textbf{DexForce}~\cite{dexforce2025}, which generates force-informed fingertip targets from one successful demonstration per task and tracks them with Cartesian impedance. The adaptation retains arm replay, uses different demonstration and initialization conditions. \textbf{Force Tracking} retains $L_F$, residual regularization, and bounds with $w_\tau=0$, while \textbf{DexTeleop-0} adds moment regularization. Controller parameters are held fixed across tasks within each hand configuration, with no task-specific tuning.}

\subsubsection{Task Specifications and Performance Evaluation}\label{subsec:metrics}
\textcolor{black}{Four protocols evaluate simulation replay, hardware replay under identical nominal commands, the User Study, and applied external disturbances. Initial object-position offsets in hardware replay are distinct from forces applied during a grasp.}

\textit{Simulation replay:} Single-arm Ball Assembly and bimanual Stir in Cup reuse recorded nominal commands for \textcolor{black}{100 trials per method for each task}, varying mass and friction within Table~\ref{tab:simulation_randomization_ranges}. \textcolor{black}{Force mean, sample SD, and P95 summarize right-thumb trial-average magnitudes over positive-force frames, including failed trials. P95 denotes the 95th percentile, computed by Type 7 linear interpolation. Positive-frame percentage measures contact coverage. Stir success concerns lift and insertion.}

\input{revision_shared/range_sim_tests.tex}

\begin{table}[!htb]
\caption{Simulation replay comparison \textcolor{black}{(100 trials per method per task)}.}\label{tab:simulation_results}
  \begingroup
  \centering
  \setlength{\tabcolsep}{3pt}%
  \begin{adjustbox}{max width=\linewidth}
    \input{revision_shared/simulation_table.tex}  \end{adjustbox}\par
  \endgroup

\par\smallskip\footnotesize \textcolor{black}{Forces use trial-average positive-frame thumb magnitudes (N), as defined in Section~\ref{subsec:metrics}. Stir Stages 1 and 2 are scored independently.}
\end{table}

\textcolor{black}{\textit{Identical-command hardware comparison:} Each method receives the same recorded nominal commands within a given hand and task, with corrections recomputed under initial object-placement variation. Sharpa and Wuji use hand-specific recordings for prism placement, food placement, and bimanual pouring. Table~\ref{tab:replay} reports ten trials per method for each condition.}

\textcolor{black}{\textit{User Study:} Five novices and two experienced operators perform single-arm Gear Mesh and Peg Insertion and bimanual Food Sorting and Tube Operation. Each makes five attempts per task with each method, giving 25 novice trials and 10 experienced trials per method for each task. Participants draw lots to determine method order and receive a basic functional introduction without task-operation training. All methods follow common object-placement rules, with normal manual position and orientation variation. Gear Mesh and Peg Insertion have two stages, Food Sorting four, and Tube Operation three. Stage rates use all group attempts, while full-task success requires every stage.}

\textcolor{black}{\textit{External-force disturbance:} All methods use the same fixed-arm Sharpa hand configuration to grasp a spherical watermelon model, with a consistent manual disturbance applied to one half. The supplementary video shows the procedure, which has normal amplitude and timing variation from manual application.}

\subsection{Simulation Results Analysis}
\label{subsec:comparison_results}\label{subsec:sim_results}

\begin{figure}[!htb]
\centering
\includegraphics[width=0.70\columnwidth,pagebox=cropbox]{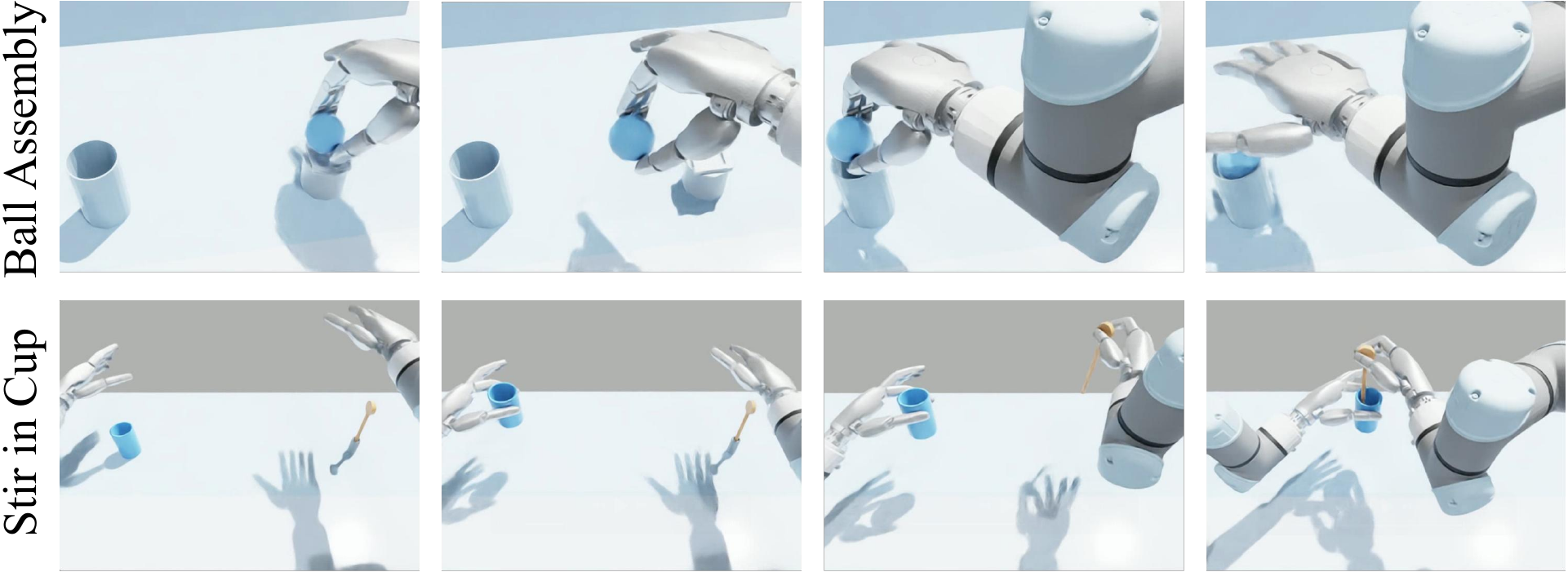}
\caption{Simulation tasks: Ball Assembly (top) and Stir in Cup (bottom), with four snapshots per row. \textcolor{black}{Stir scoring concerns lift and insertion events.}}
\label{fig:sim_experiments}
\end{figure}
\textcolor{black}{Controlled replay provides common nominal commands for the tasks in Fig.~\ref{fig:sim_experiments}, enabling comparisons across mass and friction variation. Table~\ref{tab:simulation_results} shows the highest final-stage success for DexTeleop-0, with 91\% in Ball Assembly and 62\% in Stir in Cup, compared with 88\% and 49\% for Force Tracking. The larger Stir gain indicates task dependence. Initial Ball success is slightly lower than for No Residual and Linear. DexForce has lower mean forces but lower completion and contact coverage, showing why force magnitude alone is insufficient to characterize manipulation performance.}

\textcolor{black}{A plausible explanation lies in the contact objectives. The stiffness--Jacobian model maps separate contact-force errors to bounded joint corrections, limiting departures from nominal motion. Unlike independent Linear adjustment, the optimization considers the effects of one hand's corrections on all active contacts. The moment term can additionally coordinate contacts that produce rotational imbalance about the contact centroid. This provides a mechanism-based interpretation of the task-level trend, with its effect depending on active contacts and the predicted-moment deadband.}

\begin{figure}[!htb]
\centering
\begingroup
\captionsetup[sub]{skip=2pt}
\sbox{\contactResultantPanel}{\subcaptionbox{\textcolor{black}{Contact-force resultant magnitude.}\label{fig:trial_resultant}}{\includegraphics[width=.48\linewidth]{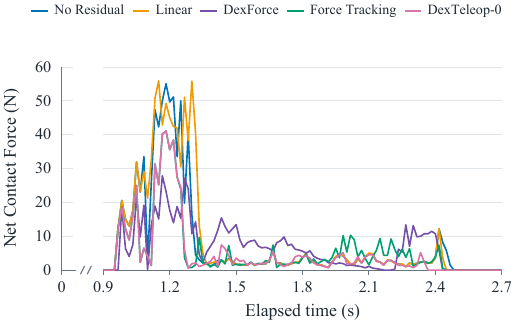}}}
\sbox{\contactForcesPanel}{\subcaptionbox{\textcolor{black}{Individual contact-force magnitudes.}\label{fig:trial_forces}}{\includegraphics[width=.48\linewidth]{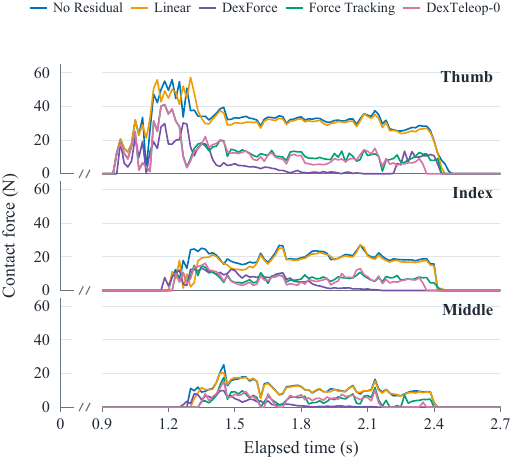}}}
\sbox{\contactResidualsPanel}{\subcaptionbox{\textcolor{black}{Thumb and index joint residuals.}\label{fig:trial_residuals}}{\includegraphics[width=.48\linewidth]{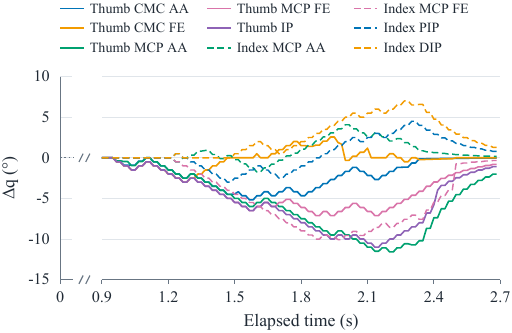}}}
\begin{minipage}[t]{.48\linewidth}
\vspace{0pt}\usebox{\contactResultantPanel}
\par\medskip
\usebox{\contactResidualsPanel}
\end{minipage}\hfill
\begin{minipage}[t]{.48\linewidth}
\vspace{0pt}\usebox{\contactForcesPanel}
\end{minipage}
\endgroup
\caption{\textcolor{black}{Contact behavior in Ball Assembly simulation replay under common recorded nominal commands. (a) Magnitude of the vector sum of fingertip forces (N). (b) Thumb, index, and middle force magnitudes (N) for No Residual, Linear, DexForce, Force Tracking, and DexTeleop-0. (c) Joint residuals $\Delta q$ ($^\circ$) for DexTeleop-0, with five thumb joints (solid) and four index joints (dashed). Panels share elapsed time (s), omitting 0--0.9 s. After the initial transient, a small resultant coexists with substantial individual forces, particularly for No Residual and Linear.}}
\label{fig:contact_signals}\label{fig:tactile_force_analysis}\label{fig:joint_residuals}
\end{figure}

\textcolor{black}{Fig.~\ref{fig:contact_signals} illustrates the contact behavior. A small resultant in (a) coexists with substantial individual forces in (b), particularly for No Residual and Linear after the initial transient. DexTeleop-0 and Force Tracking generally maintain lower individual loads, with differences across fingers and time. This motivates separate per-contact errors, since opposing forces can cancel in the vector sum. Panel (c) shows joint corrections on the same time axis. These profiles complement success and contact coverage when interpreting the controller response.}

\begin{figure}[!htb]
\centering
\begingroup
\captionsetup[sub]{skip=2pt}
\begin{subfigure}[t]{.49\linewidth}
\centering
\includegraphics[width=\linewidth,pagebox=cropbox]{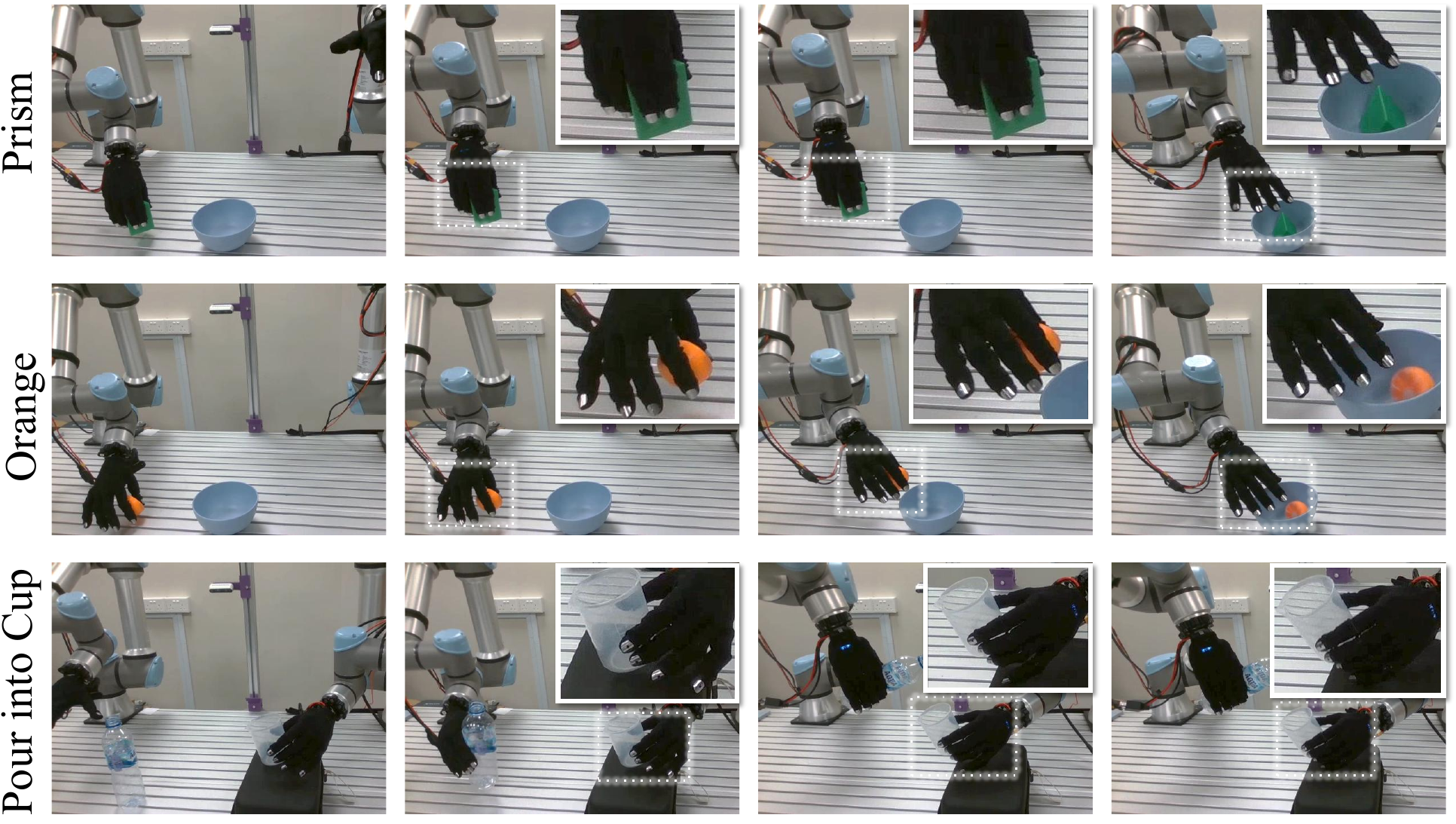}
\caption{\textcolor{black}{Task execution with Wuji Hand 2.}}\label{fig:replay_wuji}
\end{subfigure}\hfill
\begin{subfigure}[t]{.49\linewidth}
\centering
\includegraphics[width=\linewidth,pagebox=cropbox]{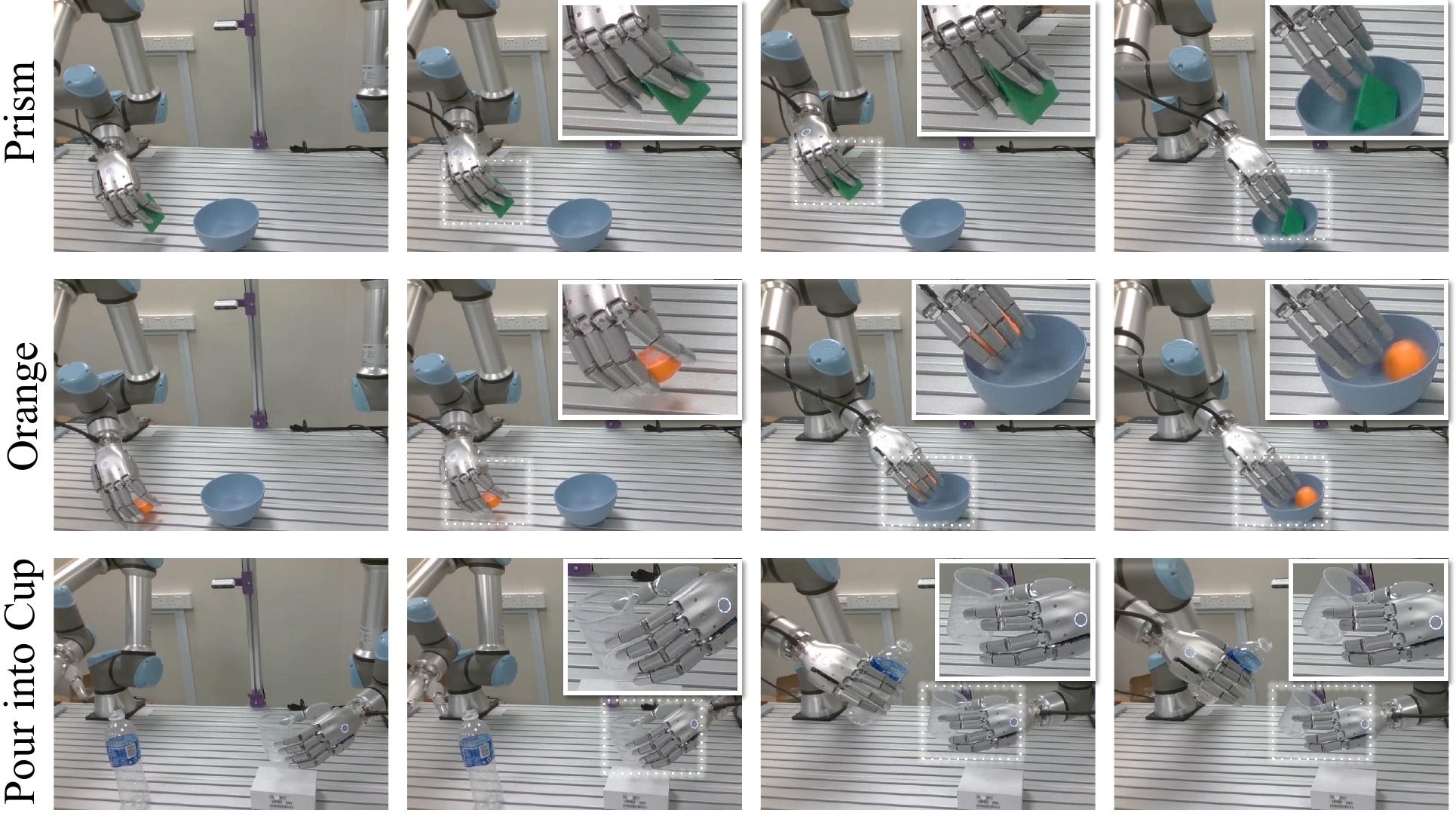}
\caption{\textcolor{black}{Task execution with Sharpa Wave.}}\label{fig:replay_sharpa}
\end{subfigure}
\endgroup
\caption{\textcolor{black}{Cross-platform demonstration of DexTeleop-0 on (a) Wuji Hand 2 and (b) Sharpa Wave for prism placement, orange placement, and bimanual pouring, using hand-specific sensing configurations and identical nominal control inputs. Insets show enlarged views of hand–object interactions.}}
\label{fig:replay}
\end{figure}

\begin{table}[!htb]
\caption{\textcolor{black}{Hardware comparison under identical nominal commands: success rate in ten trials (\%).}}\label{tab:replay}
  \begingroup
  \centering
  \setlength{\tabcolsep}{3pt}%
  \begin{adjustbox}{max width=\linewidth}
    \input{revision_shared/replay_table.tex}  \end{adjustbox}\par
  \endgroup

\par\smallskip\footnotesize \textcolor{black}{Bold denotes the largest displayed success rate for each hand and task.}
\end{table}

\Needspace{7\baselineskip}
\subsection{Real-world Results Analysis}\label{subsec:hardware_results}
\subsubsection{\textcolor{black}{Comparison Under Identical Nominal Commands}}\label{subsec:replay_results}\label{subsec:targeted}
\textcolor{black}{For the tasks in Fig.~\ref{fig:replay}, DexTeleop-0 achieves the highest displayed success in all six conditions (Table~\ref{tab:replay}), with smaller gains for Sharpa pouring. With nominal commands fixed within each condition, the results illustrate contact correction under object-placement variation using direct tactile feedback or current-based force estimates.}

\Needspace{5\baselineskip}
\subsubsection{\textcolor{black}{User Study}}\label{subsec:human_results}
\textcolor{black}{Force statistics cover complete operation records, including successful and failed processes, so differences in task progression and contact duration mean that contact conditions are not strictly matched. They serve as supporting evidence alongside success rates, with DexTeleop-0's force magnitudes remaining in a comparable range to those of the comparison methods across the four tasks.}
\begin{figure}[H]
\centering
\includegraphics[width=.70\linewidth,pagebox=cropbox]{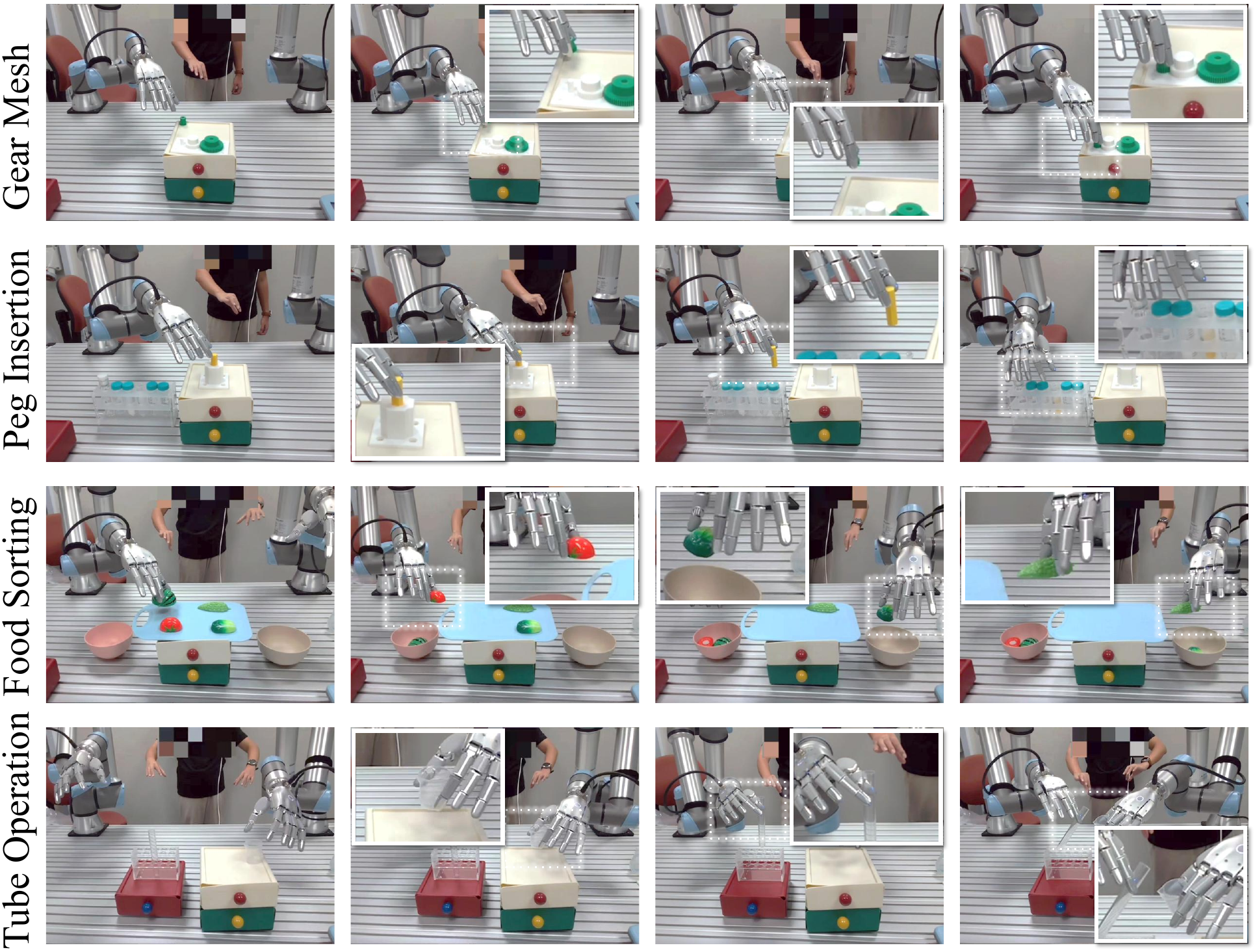}
\caption{\textcolor{black}{User Study tasks}, top to bottom: Gear Mesh, Peg Insertion, Food Sorting, and Tube Operation. Insets enlarge contact regions.}
\label{fig:real_world_experiments}\label{fig:task_examples}
\end{figure}

The four tasks in Fig.~\ref{fig:task_examples} cover single-arm assembly and bimanual coordination. \textcolor{black}{Table~\ref{tab:gear_peg_merged} separates initial acquisition from the subsequent contact-rich stage. In Gear Mesh, DexTeleop-0 increases insertion success over Force Tracking from 28\% to 44\% for novices and from 80\% to 90\% for experienced operators. Peg Insertion shows a different pattern: novice pick success reaches 100\%, but insertion success is 52\%, below Force Tracking's 64\%. Experienced operators instead improve from 60\% to 80\% at insertion. For novice Peg Insertion, the lower mean thumb force with DexTeleop-0 (3.57 versus 3.86~N) accompanies lower insertion success, illustrating why force reduction and task completion must be assessed jointly. Thus successful acquisition does not ensure successful alignment and engagement. A plausible interpretation is that local contact corrections assist some execution errors while leaving task geometry and the operator's nominal approach important; the controller adjusts hand joints while retaining nominal arm commands.}

\begin{table}[H]
\caption{\textcolor{black}{User Study}: Gear Mesh and Peg Insertion stages and thumb force.}\label{tab:gear_peg_merged}
  \begingroup
  \centering
  \setlength{\tabcolsep}{3pt}%
  \begin{adjustbox}{max width=\linewidth}
    \input{revision_shared/gear_peg_table.tex}  \end{adjustbox}\par
  \endgroup

\end{table}

\textcolor{black}{The bimanual results also depend on task and experience. Food Sorting final-stage success rises from 48\% to 68\% for novices and from 80\% to 90\% for experienced operators relative to Force Tracking (Table~\ref{tab:food_sorting}). Tube Operation improves from 52\% to 76\% and from 70\% to 80\%, respectively (Table~\ref{tab:tube_operation}). These are stage outcomes over all group attempts, rather than conditional probabilities of completing the remaining sequence. Full-task success requires every stage: novice Tube Operation has 72\% full-task success despite 76\% final-stage success. With 25 novice and 10 experienced attempts per method and task, one completion changes the displayed rate by four or ten percentage points. The group comparisons therefore describe the recorded task-dependent pattern without establishing a uniform benefit for every operator or manipulation stage.}

\begin{table}[H]
\caption{\textcolor{black}{User Study}: Food Sorting.}\label{tab:food_sorting}
  \begingroup
  \centering
  \setlength{\tabcolsep}{3pt}%
  \begin{adjustbox}{max width=\linewidth}
    \input{revision_shared/food_table.tex}  \end{adjustbox}\par
  \endgroup

\end{table}

\clearpage
\begin{table}[H]
\caption{\textcolor{black}{User Study}: Tube Operation.}\label{tab:tube_operation}
  \begingroup
  \centering
  \setlength{\tabcolsep}{3pt}%
  \begin{adjustbox}{max width=\linewidth}
    \input{revision_shared/tube_table.tex}  \end{adjustbox}\par
  \endgroup

\end{table}
\begin{figure}[H]
\centering
\includegraphics[width=.85\linewidth]{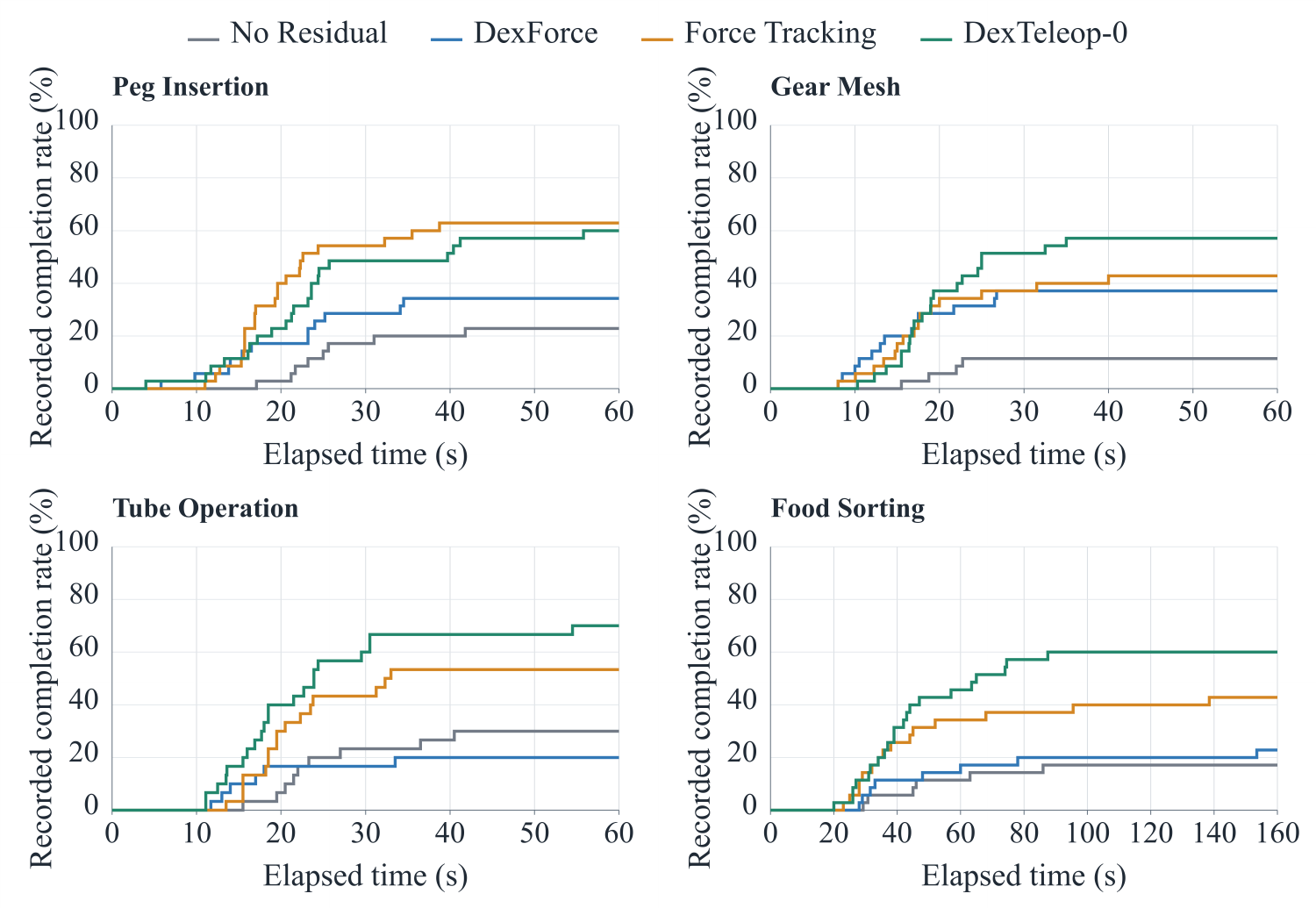}
\caption{\textcolor{black}{Recorded completion rate versus elapsed time for No Residual, DexForce, Force Tracking, and DexTeleop-0. Each curve reports the fraction of recorded attempts that completed the task by the indicated time, including failed attempts in the denominator.}}
\label{fig:real_completion_curves}
\end{figure}

\textcolor{black}{Fig.~\ref{fig:real_completion_curves} compares recorded completion rates over time. DexTeleop-0 leads at later displayed times in Gear Mesh, Tube Operation, and Food Sorting, while Force Tracking leads in Peg Insertion. The curves reflect both completed attempts and their execution time.}

\Needspace{5\baselineskip}
\subsubsection{\textcolor{black}{External-force disturbance experiments}}\label{subsec:disturbance_results}
\begin{figure}[H]
\centering
\includegraphics[width=\columnwidth]{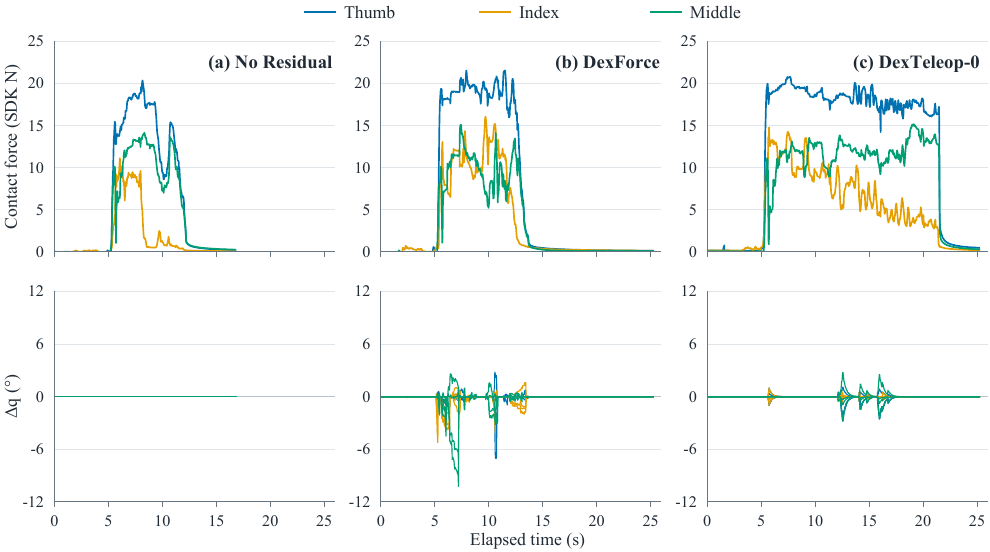}
\caption{\textcolor{black}{Contact response to manual disturbance of the watermelon-model grasp. Columns compare No Residual, DexForce, and DexTeleop-0. Top: finger forces (N). Bottom: joint residuals ($^\circ$).}}
\label{fig:external_disturbance}
\end{figure}

\textcolor{black}{Fig.~\ref{fig:external_disturbance} compares the fixed-arm grasp under the common manual-disturbance procedure. No Residual and DexForce forces decay near zero at approximately 12--14~s, whereas DexTeleop-0 retains thumb and middle contact until about 21~s. The different finger-force trajectories show why contact persistence and load distribution should be considered together when assessing the maintained grasp.}

\textcolor{black}{The residual traces distinguish passive tracking from corrective action. No Residual has zero joint residual by construction but loses contact, so small corrections alone do not indicate successful disturbance handling. DexTeleop-0 instead combines persistent contact with smaller visible residual excursions than DexForce. This is consistent with the stiffness--Jacobian model in (\ref{eq:force_model}), through which bounded hand-joint corrections adapt the commanded posture as measured contact loading changes.}

\textcolor{black}{The comparison remains qualitative: manual forcing has normal amplitude and timing variation, and the applied load is not quantified in these traces. They therefore establish neither a calibrated rejection threshold nor a matched-load recovery time. Moreover, Fig.~\ref{fig:external_disturbance} omits the Force Tracking ablation, preventing attribution of the response specifically to moment regularization. The evidence concerns contact persistence and corrective motion in the displayed grasp, complementing replay tests of initial placement variation.}

\section{Conclusion}\label{sec:conclusion}
In this work, we introduced a tactile-driven adaptation strategy for bimanual dexterous teleoperation, \textcolor{black}{using fingertip feedback to refine motion commands during contact.} Instantiated within our full-stack framework, DexTeleop-0, this strategy integrates accessible egocentric tracking from a commercial VR headset with a real-time \textcolor{black}{bounded residual optimization loop}, coordinating 56 concurrent DoFs across dual arm-hand assemblies. \textcolor{black}{The formulation combines per-contact vector force tracking with predicted contact-moment regularization under joint and residual limits. Arm commands remain nominal, and each hand coordinates its contacts through a separate optimization. Simulation and physical experiments show improvements in task completion and contact persistence, with task-dependent force and timing tradeoffs. Replay under common nominal commands provides a controlled comparison of feedback corrections, while the User Study evaluates assistance during live teleoperation by novice and experienced operators. These results should be read jointly with force measurements, since successful and failed attempts can involve different contact durations and manipulation stages. The local contact model depends on sensing and stiffness estimates, and its moment penalty acts when the predicted contact moment exceeds the configured tolerance.} Future work will focus on predictive slip detection to adapt to unexpected object properties and on using visual-tactile demonstrations to train generalizable imitation learning policies.
\Needspace{4\baselineskip}
\bibliographystyle{ieeetr}
\bibliography{refrence_r1}
\end{document}

%% file: revision_shared/controller_r1.tex
\subsubsection{Localized Force Feedback Tracking}
\label{subsec:force_model}
\textcolor{black}{For one hand, $\Delta q\in\mathbb R^n$ is a joint-position residual in radians. Let $\mathcal C$ contain $m$ valid contacts. Express contact positions $p_i$ (m), forces $f_i$ (N), and Jacobians in the same hand-base frame, with $f_i$ defined as force exerted by the hand on the object. Wuji uses empirical $\mathrm N_{\rm est}$, with corresponding stiffness, moment, and normalization units. An object-on-hand measurement is negated before use. A sensor-frame force uses the configured sensor-to-base rotation, and stiffness transforms as $K_i^b=R_iK_i^sR_i^\top$. Positions additionally require translation. Kinematics are linearized at the latest nominal target.}

The controller regulates contact through position corrections rather than direct torque commands. \textcolor{black}{The translational Jacobian $J_i\in\mathbb R^{3\times n}$ (m/rad) predicts contact displacement. An effective contact stiffness relates this displacement to force variation, $\delta f_i=K_i\delta p_i$, giving the local approximation}
\begingroup\color{black}
\begin{equation}
\delta p_i=J_i\Delta q,\quad A_i=K_iJ_i,\quad
 f_i^{\rm pred}=f_i+A_i\Delta q,
\label{eq:force_model}
\end{equation}
\endgroup
\textcolor{black}{where $K_i\in\mathbb R^{3\times3}$ (N/m) is a configured model input, giving $A_i$ units N/rad. The approximation assumes fixed stiffness and contacts and negligible object motion over one update. Its accuracy depends on contact persistence, compliance, sensing, and the consistency of the measured-force and nominal-kinematics references.}

\textcolor{black}{For configured component bounds $l_i,u_i\in\mathbb R^3$ (N), the adapter constructs a fixed target $f_i^{\rm tar}=\operatorname{clip}(f_i,l_i,u_i)$ from the current force. Define $e_i=f_i-f_i^{\rm tar}\in\mathbb R^3$ and $\alpha_i=w_Lc_i/s_F^2$, where $c_i\in[0,1]$ is contact confidence, $w_L\geq0$ is a local weight, and $s_F>0$ is a force scale (N). The local cost is}
\begingroup\color{black}
\begin{equation}
 L_F(\Delta q)=\frac12\sum_{i\in\mathcal C}\alpha_i
 \|e_i+A_i\Delta q\|_2^2.
\label{eq:local_cost}
\end{equation}
\endgroup
\textcolor{black}{Clipping acts componentwise: an above-bound component targets its upper bound, a below-bound component targets its lower bound, and an in-range component targets its current value. Thus $e_i+A_i\Delta q=f_i^{\rm pred}-f_i^{\rm tar}$ is the predicted force error. Minimizing it favors a kinematically feasible displacement that opposes the error through $K_iJ_i$, rather than assigning a fixed correction sign to each joint. The correction enters the next hand command as $q_{\rm cmd}=q_{\rm nom}+\Delta q$. The $3m$ contact residuals remain separate, preventing cancellation across fingers. Targets are fixed during the solve: the cost penalizes vector changes rather than only predicted violations of the force interval. Confidence weights reduce the influence of less reliable contacts.}

\subsubsection{Multi-Contact Force--Torque Balance Optimization}
\label{subsec:balance}
\textcolor{black}{Let $r_0=m^{-1}\sum_{i\in\mathcal C}p_i$ be the current contact centroid, frozen within the update as a geometric moment reference. The contact resultant and moment are}
\begingroup\color{black}
\begin{equation}
 F=\sum_{i\in\mathcal C}f_i,\qquad
 \tau=\sum_{i\in\mathcal C}(p_i-r_0)\times f_i.
\label{eq:resultants}
\end{equation}
\endgroup
\textcolor{black}{$F$ (N) and $\tau$ (N\,m) describe the same contacts but capture different effects: force changes that partly cancel in $F$ can still change the moment through their different lever arms. Quasi-static object balance requires these resultants and all external loads to sum to zero. The present model regulates hand contacts, with object weight and unmeasured environment contacts outside its scope. For fixed contacts, $F^{\rm pred}=F+\sum_i A_i\Delta q$. The moment also changes when contact points move: expanding $(p_i+J_i\Delta q-r_0)\times(f_i+A_i\Delta q)$ and discarding terms quadratic in $\Delta q$ gives $\delta\tau=B_\tau\Delta q$. With $[v]_\times u=v\times u$,}
\begingroup\color{black}
\begin{equation}
 B_\tau=\sum_{i\in\mathcal C}
 \left([p_i-r_0]_\times A_i-[f_i]_\times J_i\right),
\label{eq:torque_jacobian}
\end{equation}
\endgroup
\textcolor{black}{where $B_\tau\in\mathbb R^{3\times n}$ has units N\,m/rad. The second term accounts for moving contact points. Define $e_\tau=\tau+\tau_{\rm ext}-\tau^{\rm tar}$, where $\tau^{\rm tar}$ is the configured moment target and $\tau_{\rm ext}$ a supplied external-moment estimate (N\,m). A zero nominal gravity moment treats gravity as an unmodeled load when object weight and center of mass are unknown.}

\subsubsection{Unified Force Balance Optimization}
\label{subsec:residual_optimization}
\textcolor{black}{The contact estimator determines which measurements enter $L_F$ and the contact moment. The same $\Delta q$ couples their predicted changes. As illustrated in Fig.~\ref{fig:tactile_balancing_principle}, force regulation changes individual contact loading, while the moment term discourages their combined rotational imbalance through a soft local objective. With nonnegative slack $z\in\mathbb R^3$ (N\,m), the residual QP is}
\begingroup\color{black}
\begin{equation}
\begin{aligned}
\min_{\Delta q,z}\quad& L_F(\Delta q)+\frac{\lambda}{2}\|\Delta q\|_2^2
             +\frac{w_\tau}{2s_\tau^2}\|z\|_2^2\\
\text{s.t.}\quad&-d_\tau\mathbf1-z\leq e_\tau+B_\tau\Delta q
                             \leq d_\tau\mathbf1+z,\\
&z\geq0,\quad \Delta q\in\mathcal Q_t.
\end{aligned}
\label{eq:residual_qp}
\end{equation}
\endgroup
\textcolor{black}{Here $\lambda>0$ regularizes joint residuals, $w_\tau\geq0$ is the moment weight, $s_\tau>0$ the moment scale (N\,m), and $d_\tau\geq0$ the per-axis moment tolerance (N\,m). The force and moment scales normalize quantities with different physical units, while regularization favors small departures from the nominal command. Eliminating $z$ gives $\sum_a\max(|e_{\tau,a}+(B_\tau\Delta q)_a|-d_\tau,0)^2$. The deadband applies to the \emph{predicted} moment. With fixed contacts, this is a convex QP, and $\lambda>0$ yields a unique joint residual whenever the feasible set is nonempty.}

\textcolor{black}{The feasible set $\mathcal Q_t$ intersects physical joint limits, residual amplitude bounds, and consecutive residual-change bounds:}
\begingroup
\begin{equation}
\begin{aligned}
q_{\min}-q_{\rm nom}&\leq\Delta q\leq q_{\max}-q_{\rm nom},\\
\textcolor{black}{-\bar q\leq\Delta q}&\textcolor{black}{\leq\bar q},\qquad
-\delta\leq\Delta q-\Delta q_{\rm prev}\leq\delta.
\end{aligned}
\label{eq:bounds}
\end{equation}
\endgroup
\textcolor{black}{The last bound limits changes in the residual component of the hand command.}

\textcolor{black}{The local quadratic is defined by $H_L=\sum_i\alpha_iA_i^\top A_i+\lambda I$ and $g_L=\sum_i\alpha_iA_i^\top e_i$, incorporating local tracking and residual regularization once each. Net force is used as a diagnostic quantity, while the objective combines the per-contact force errors and contact moment. Force Tracking sets $w_\tau=0$, whereas the full method enables moment regularization with sufficient valid contacts. Solutions are checked for numerical validity, with local-force fallback in the Wuji adapter.}

\textcolor{black}{Contact-set changes can shift $r_0$ between updates: moving the reference by $d$ changes moment to $\tau'=\tau-d\times F$. Each hand regulates its own contact moment about the current centroid, while the nominal teleoperation commands coordinate the bimanual motion.}

%% file: revision_shared/range_sim_tests.tex
\begin{table}[!t]
\centering
\caption{\textcolor{black}{Physical parameter randomization ranges in the simulation experiments.}}
\label{tab:simulation_randomization_ranges}
\begingroup
\small
\renewcommand{\arraystretch}{1.3}
\setlength{\tabcolsep}{3pt}
\begin{tabular*}{\linewidth}{@{\extracolsep{\fill}}llccc@{}}
\toprule
\textcolor{black}{\textbf{Task}} &
\textcolor{black}{\textbf{Object}} &
\textcolor{black}{\textbf{Mass (kg)}} &
\textcolor{black}{\shortstack{\textbf{SF}\,$\mu_s$}} &
\textcolor{black}{\shortstack{\textbf{DF}\,$\mu_d$}} \\
\midrule
\textcolor{black}{Ball Assembly} &
\textcolor{black}{Ball} &
\textcolor{black}{0.190--0.230} &
\textcolor{black}{2.5--2.9} &
\textcolor{black}{2.4--2.7} \\
\midrule
\multirow{2}{*}{\textcolor{black}{Stir in Cup}} &
\textcolor{black}{Cup} &
\textcolor{black}{0.158--0.162} &
\textcolor{black}{2.2--2.5} &
\textcolor{black}{1.8--2.1} \\
&
\textcolor{black}{Rod} &
\textcolor{black}{0.023--0.025} &
\textcolor{black}{1.9--2.2} &
\textcolor{black}{1.7--2.0} \\
\bottomrule
\end{tabular*}
\par\vspace{3pt}
\begin{minipage}{\linewidth}
\footnotesize
\textcolor{black}{Entries give the lower and upper bounds for each
randomized object property. Static and dynamic friction coefficients (SF and DF in the table)
are dimensionless.}
\end{minipage}
\endgroup
\end{table}

%% file: revision_shared/simulation_table.tex
\begin{tabular}{@{}lcccccc@{}}
\toprule
\textbf{Method} & \shortstack{\textbf{Stage}\\\textbf{1}} & \shortstack{\textbf{Stage}\\\textbf{2}} & \shortstack{\textbf{Stage}\\\textbf{3}} & \textcolor{black}{\shortstack{\textbf{Contact force}\\\textbf{mean $\pm$ SD (N)}}} & \textcolor{black}{\textbf{P95 (N)}} & \textcolor{black}{\shortstack{\textbf{Positive}\\\textbf{frames (\%)}}} \\
\midrule
\multicolumn{7}{@{}l}{\textbf{Ball Assembly}} \\
No Residual & 99\%  & \textcolor{black}{13\% } & \textcolor{black}{11\% } & \textcolor{black}{$30.48 \pm 4.18$} & \textcolor{black}{31.99} & \textcolor{black}{54.57\%} \\
\textcolor{black}{Linear} & 100\% & \textcolor{black}{5\%  } & 1\%  & \textcolor{black}{$30.82 \pm 1.43$} & \textcolor{black}{31.94} & \textcolor{black}{55.73\%} \\
\textcolor{black}{DexForce} & \textcolor{black}{100\%} & \textcolor{black}{39\%} & \textcolor{black}{38\%} & \textcolor{black}{$11.49 \pm 3.26$} & \textcolor{black}{15.98} & \textcolor{black}{49.33\%} \\
Force Tracking & \textcolor{black}{97\% } & \textcolor{black}{88\% } & \textcolor{black}{88\%} & \textcolor{black}{$13.83 \pm 1.45$} & \textcolor{black}{14.77} & \textcolor{black}{51.27\%} \\
DexTeleop-0 & \textcolor{black}{97\%} & \textcolor{black}{91\% } & \textcolor{black}{91\%} & \textcolor{black}{$13.29 \pm 1.21$} & \textcolor{black}{14.41} & \textcolor{black}{51.65\%} \\
\midrule
\multicolumn{7}{@{}l}{\textbf{Stir in Cup}} \\
No Residual & 100\% & 100\% & \textcolor{black}{49\%} & \textcolor{black}{$3.94 \pm 0.89$} & \textcolor{black}{5.17} & \textcolor{black}{40.60\%} \\
\textcolor{black}{Linear} & 100\% & 100\% & \textcolor{black}{44\%} & \textcolor{black}{$3.96 \pm 0.83$} & \textcolor{black}{4.98} & \textcolor{black}{37.24\%} \\
\textcolor{black}{DexForce} & \textcolor{black}{95\%} & \textcolor{black}{100\%} & \textcolor{black}{3\%} & \textcolor{black}{$2.50 \pm 0.25$} & \textcolor{black}{2.75} & \textcolor{black}{10.58\%} \\
Force Tracking & 100\% & 100\% & 49\% & \textcolor{black}{$3.96 \pm 0.89$} & \textcolor{black}{5.16} & \textcolor{black}{39.74\%} \\
DexTeleop-0 & 100\% & 100\% & \textcolor{black}{\textbf{62\%}} & \textcolor{black}{$3.85 \pm 0.81$} & \textcolor{black}{5.11} & \textcolor{black}{41.56\%} \\
\bottomrule
\end{tabular}

%% file: revision_shared/replay_table.tex
\begingroup\color{black}
\begin{tabular}{llccc}
\toprule
Hand & Task & No Residual & DexForce & DexTeleop-0 \\
\midrule
\multirow{3}{*}{Wuji} & Food placement & 40\% (4/10) & 30\% (3/10) & \textbf{80\% (8/10)} \\
& Triangular prism & 40\% (4/10) & 50\% (5/10) & \textbf{90\% (9/10)} \\
& Bimanual pouring & 0\% (0/10) & 10\% (1/10) & \textbf{40\% (4/10)} \\
\midrule
\multirow{3}{*}{Sharpa} & Food placement & 30\% (3/10) & 40\% (4/10) & \textbf{90\% (9/10)} \\
& Triangular prism & 50\% (5/10) & 40\% (4/10) & \textbf{80\% (8/10)} \\
& Bimanual pouring & 40\% (4/10) & 50\% (5/10) & \textbf{60\% (6/10)} \\
\bottomrule
\end{tabular}

\endgroup

%% file: revision_shared/gear_peg_table.tex
\begin{tabular}{llcccc}
\toprule
\textbf{Scene} & \textbf{Method} & \textbf{Pick Stage} & \textbf{Insert Stage} & \textcolor{black}{\shortstack{\textbf{Thumb force}\\mean $\pm$ SD \textbf{(N)}}} & \textcolor{black}{\shortstack{\textbf{P95}\\\textbf{(N)}}} \\
\midrule
\multicolumn{6}{l}{\textcolor{black}{Novice operators}} \\
\midrule
\multirow{4}{*}{Gear Mesh} & No Residual & \textcolor{black}{56\%} & \textcolor{black}{12\%} & \textcolor{black}{$2.76 \pm 3.41$} & \textcolor{black}{10.98} \\
 & \textcolor{black}{DexForce} & \textcolor{black}{72\%} & \textcolor{black}{20\%} & \textcolor{black}{$3.05 \pm 3.36$} & \textcolor{black}{11.93} \\
 & Force Tracking & \textcolor{black}{92\%} & \textcolor{black}{28\%} & \textcolor{black}{$2.16 \pm 1.70$} & \textcolor{black}{5.80} \\
 & DexTeleop-0 & \textcolor{black}{96\%} & \textcolor{black}{44\%} & \textcolor{black}{$3.15 \pm 2.21$} & \textcolor{black}{7.36} \\
\midrule
\multirow{4}{*}{Peg Ins.} & No Residual & \textcolor{black}{72\%} & \textcolor{black}{28\%} & \textcolor{black}{$5.11 \pm 5.14$} & \textcolor{black}{14.34} \\
 & \textcolor{black}{DexForce} & \textcolor{black}{72\%} & \textcolor{black}{24\%} & \textcolor{black}{$6.58 \pm 7.33$} & \textcolor{black}{21.05} \\
 & Force Tracking & \textcolor{black}{92\%} & \textcolor{black}{64\%} & \textcolor{black}{$3.86 \pm 2.53$} & \textcolor{black}{8.16} \\
 & DexTeleop-0 & \textcolor{black}{100\%} & \textcolor{black}{52\%} & \textcolor{black}{$3.57 \pm 2.53$} & \textcolor{black}{8.04} \\
\midrule
\multicolumn{6}{l}{\textcolor{black}{Experienced operators}} \\
\midrule
\multirow{4}{*}{Gear Mesh} & No Residual & \textcolor{black}{80\%} & \textcolor{black}{10\%} & \textcolor{black}{$2.31 \pm 1.50$} & \textcolor{black}{5.13} \\
 & \textcolor{black}{DexForce} & \textcolor{black}{90\%} & \textcolor{black}{80\%} & \textcolor{black}{$3.10 \pm 2.26$} & \textcolor{black}{7.15} \\
 & Force Tracking & \textcolor{black}{100\%} & \textcolor{black}{80\%} & \textcolor{black}{$2.28 \pm 2.17$} & \textcolor{black}{6.51} \\
 & DexTeleop-0 & \textcolor{black}{100\%} & \textcolor{black}{90\%} & \textcolor{black}{$1.95 \pm 1.79$} & \textcolor{black}{5.69} \\
\midrule
\multirow{4}{*}{Peg Ins.} & No Residual & \textcolor{black}{80\%} & \textcolor{black}{20\%} & \textcolor{black}{$3.58 \pm 5.46$} & \textcolor{black}{16.21} \\
 & \textcolor{black}{DexForce} & \textcolor{black}{100\%} & \textcolor{black}{60\%} & \textcolor{black}{$3.36 \pm 3.48$} & \textcolor{black}{10.83} \\
 & Force Tracking & \textcolor{black}{90\%} & \textcolor{black}{60\%} & \textcolor{black}{$3.26 \pm 2.00$} & \textcolor{black}{6.79} \\
 & DexTeleop-0 & \textcolor{black}{90\%} & \textcolor{black}{80\%} & \textcolor{black}{$4.66 \pm 2.46$} & \textcolor{black}{9.29} \\
\bottomrule
\end{tabular}

%% file: revision_shared/food_table.tex
\begin{tabular}{lcccccc}
\toprule
\textbf{Method} & \textbf{Stage 1} & \textbf{Stage 2} & \textbf{Stage 3} & \textbf{Stage 4} & \textcolor{black}{\shortstack{\textbf{Thumb force}\\mean $\pm$ SD \textbf{(N)}}} & \textcolor{black}{\shortstack{\textbf{P95}\\\textbf{(N)}}} \\
\midrule
\multicolumn{7}{l}{\textcolor{black}{Novice Operators}} \\
\midrule
No Residual & \textcolor{black}{60\%} & \textcolor{black}{44\%} & \textcolor{black}{56\%} & \textcolor{black}{48\%} & \textcolor{black}{$6.27 \pm 6.18$} & \textcolor{black}{18.43} \\
\textcolor{black}{DexForce} & \textcolor{black}{72\%} & \textcolor{black}{68\%} & \textcolor{black}{56\%} & \textcolor{black}{36\%} & \textcolor{black}{$5.18 \pm 5.15$} & \textcolor{black}{16.43} \\
Force Tracking & \textcolor{black}{84\%} & \textcolor{black}{92\%} & \textcolor{black}{84\%} & \textcolor{black}{48\%} & \textcolor{black}{$5.28 \pm 3.95$} & \textcolor{black}{13.09} \\
DexTeleop-0 & \textcolor{black}{92\%} & \textcolor{black}{96\%} & \textcolor{black}{80\%} & \textcolor{black}{68\%} & \textcolor{black}{$5.13 \pm 3.92$} & \textcolor{black}{13.53} \\
\midrule
\multicolumn{7}{l}{\textcolor{black}{Experienced Operators}} \\
\midrule
No Residual & \textcolor{black}{100\%} & \textcolor{black}{70\%} & \textcolor{black}{50\%} & \textcolor{black}{50\%} & \textcolor{black}{$6.38 \pm 5.63$} & \textcolor{black}{18.50} \\
\textcolor{black}{DexForce} & \textcolor{black}{50\%} & \textcolor{black}{70\%} & \textcolor{black}{80\%} & \textcolor{black}{90\%} & \textcolor{black}{$4.52 \pm 4.68$} & \textcolor{black}{14.83} \\
Force Tracking & \textcolor{black}{80\%} & \textcolor{black}{100\%} & \textcolor{black}{90\%} & \textcolor{black}{80\%} & \textcolor{black}{$4.85 \pm 4.60$} & \textcolor{black}{13.14} \\
DexTeleop-0 & \textcolor{black}{90\%} & \textcolor{black}{100\%} & \textcolor{black}{90\%} & \textcolor{black}{90\%} & \textcolor{black}{$5.14 \pm 2.76$} & \textcolor{black}{10.48} \\
\bottomrule
\end{tabular}

%% file: revision_shared/tube_table.tex
\begin{tabular}{lccccc}
\toprule
\textbf{Method} & \textbf{Stage 1} & \textbf{Stage 2} & \textbf{Stage 3} & \textcolor{black}{\shortstack{\textbf{Thumb force}\\mean $\pm$ SD \textbf{(N)}}} & \textcolor{black}{\shortstack{\textbf{P95}\\\textbf{(N)}}} \\
\midrule
\multicolumn{6}{l}{\textcolor{black}{Novice operators}} \\
\midrule
No Residual & \textcolor{black}{52\%} & \textcolor{black}{68\%} & \textcolor{black}{28\%} & \textcolor{black}{$9.47 \pm 6.83$} & \textcolor{black}{20.17} \\
\textcolor{black}{DexForce} & \textcolor{black}{80\%} & \textcolor{black}{56\%} & \textcolor{black}{20\%} & \textcolor{black}{$7.33 \pm 5.54$} & \textcolor{black}{18.24} \\
Force Tracking & \textcolor{black}{92\%} & \textcolor{black}{92\%} & \textcolor{black}{52\%} & \textcolor{black}{$5.22 \pm 2.98$} & \textcolor{black}{9.65} \\
DexTeleop-0 & \textcolor{black}{92\%} & \textcolor{black}{100\%} & \textcolor{black}{76\%} & \textcolor{black}{$5.36 \pm 2.98$} & \textcolor{black}{9.72} \\
\midrule
\multicolumn{6}{l}{\textcolor{black}{Experienced operators}} \\
\midrule
No Residual & \textcolor{black}{80\%} & \textcolor{black}{80\%} & \textcolor{black}{50\%} & \textcolor{black}{$5.20 \pm 4.16$} & \textcolor{black}{12.62} \\
\textcolor{black}{DexForce} & \textcolor{black}{90\%} & \textcolor{black}{70\%} & \textcolor{black}{40\%} & \textcolor{black}{$6.98 \pm 5.04$} & \textcolor{black}{15.35} \\
Force Tracking & \textcolor{black}{90\%} & \textcolor{black}{90\%} & \textcolor{black}{70\%} & \textcolor{black}{$4.54 \pm 2.73$} & \textcolor{black}{9.19} \\
DexTeleop-0 & \textcolor{black}{100\%} & \textcolor{black}{100\%} & \textcolor{black}{80\%} & \textcolor{black}{$5.49 \pm 2.96$} & \textcolor{black}{10.12} \\
\bottomrule
\end{tabular}